\def\year{2022}\relax
\pdfoutput=1
\documentclass[letterpaper]{article} 
\usepackage{aaai22}  
\usepackage{times}  
\usepackage{helvet}  
\usepackage{courier}  
\usepackage[hyphens]{url}  
\usepackage{graphicx} 
\usepackage{natbib}  
\usepackage{caption} 
\DeclareCaptionStyle{ruled}{labelfont=normalfont,labelsep=colon,strut=off} 
\usepackage{ifthen}
\usepackage[normalem]{ulem} 
\usepackage{xcolor}

\newboolean{showedits}
\setboolean{showedits}{true} 
\ifthenelse{\boolean{showedits}}
{
	\newcommand{\del}[1]{\textcolor{red}{\sout{#1}}} 
}{
	\newcommand{\del}[1]{} 
	
}

\newboolean{showcomments}
\setboolean{showcomments}{true} 
\newcommand{\id}[1]{$-$Id: scgPaper.tex 32478 2010-04-29 09:11:32Z oscar $-$}

\ifthenelse{\boolean{showcomments}}
{\newcommand{\nbc}[3]{
		{\colorbox{#3}{\bfseries\sffamily\scriptsize\textcolor{white}{#1}}}
		{\textcolor{#3}{\sf\small$\blacktriangleright$\textit{#2}$\blacktriangleleft$}}}
	}
{\newcommand{\nbc}[3]{}
	\renewcommand{\del}[1]{} 
	}

\definecolor{ibcolor}{rgb}{0.9,0.5,0}
\definecolor{dsrcolor}{rgb}{0.5,0.6,0}
\definecolor{cfcolor}{rgb}{0,0.5,0.9}
\definecolor{oldcolor}{rgb}{0.2,0.2,0.2}
\definecolor{tdcolor}{rgb}{1.0,0,0}
\definecolor{oldcolor}{rgb}{0.5,0.5,0.5}
\definecolor{lycolor}{rgb}{0.3,0.3,0.8}
\definecolor{rfcolor}{rgb}{0.12,0.3,0.17}
\definecolor{xlcolor}{rgb}{0.8,0.3,0.3}

\newcommand{\linyun}[1]{\nbc{LY}{#1}{lycolor}}

\usepackage[switch]{lineno}
\usepackage{epsfig}
\usepackage{graphicx}
\usepackage{amsmath}
\usepackage{amssymb}
\usepackage{comment}
\usepackage[linesnumbered, ruled, vlined]{algorithm2e}
\usepackage{soul}
\usepackage{amsthm}
\usepackage{booktabs}
\usepackage{xspace}
\usepackage{xcolor}
\usepackage{multirow}
\usepackage{colortbl}
\usepackage{multicol}
\usepackage{subcaption}
\usepackage{mathtools}  
\usepackage{amsfonts}  
\usepackage{amssymb}
\usepackage{soul}
\usepackage{array}
\usepackage{enumitem}
\usepackage{bbold}

\title{DeepVisualInsight: Time-Travelling Visualization for Spatio-Temporal Causality of Deep Classification Training}

\author{
    Xianglin Yang\textsuperscript{\rm 1}\equalcontrib,
    Yun Lin\textsuperscript{\rm 1}\thanks{Corresponding author.}\equalcontrib,
    Ruofan Liu\textsuperscript{\rm 1},
    Zhenfeng He\textsuperscript{\rm 1},
    Chao Wang\textsuperscript{\rm 1},\\
    Jin Song Dong\textsuperscript{\rm 1},
    Hong Mei\textsuperscript{\rm 2}\\
}
\affiliations{
    \textsuperscript{\rm 1}School of Computing, National University of Singapore, Singapore\\
    \textsuperscript{\rm 2}Key Lab of High-Confidence Software Technology, MoE (Peking University), China\\

    xianglin@u.nus.edu, \{dcsliny, dcslirf\}@nus.edu.sg, \{he.zhenfeng, wang.chao\}@u.nus.edu, dcsdjs@nus.edu.sg, hongmei@pku.edu.cn
}

\DeclareMathOperator*{\argmax}{arg\,max}
\DeclareMathOperator{\argmin}{argmin} 
\newcommand{\tool}{DVI\xspace}
\newcommand{\fullnametool}{DeepVisualInsight\xspace}

\newtheorem{definition}{Definition}

\newcolumntype{N}{@{}m{0pt}@{}}
\newcommand{\norm}[1]{\left\lVert#1\right\rVert}

\begin{document}

\maketitle

\begin{abstract}
    Understanding how the predictions of deep learning models are formed during the training process is crucial to improve model performance and fix model defects, especially when we need to investigate nontrivial training strategies such as active learning, and track the root cause of unexpected training results such as performance degeneration.

    In this work, we propose a time-travelling visual solution \fullnametool (\tool),
    aiming to manifest the spatio-temporal causality while training a deep learning image classifier.
    The spatio-temporal causality demonstrates how the gradient-descent algorithm and various training data sampling techniques can influence and reshape the layout of learnt input representation and the classification boundaries in consecutive epochs.
    Such causality allows us to observe and analyze the whole learning process in the visible low dimensional space.
    Technically, we propose four spatial and temporal properties and design our visualization solution to satisfy them.
    These properties preserve the most important information when (inverse-)projecting input samples between the visible low-dimensional and the invisible high-dimensional space, for causal analyses.
    Our extensive experiments show that,
    comparing to baseline approaches,
    we achieve the best visualization performance regarding the spatial/temporal properties and visualization efficiency.
    Moreover, our case study shows that our visual solution can well reflect the characteristics of various training scenarios,
    showing good potential of \tool as a debugging tool for analyzing deep learning training processes.
\end{abstract}

\section{Introduction}
Interpreting model predictions is a well-reconsigned challenge when training and analyzing deep learning models \cite{zhang2021survey}.
Various explainable AI techniques have been proposed to understand model predictions including
input attribution analysis, training data analysis, model abstraction, etc.
Generally, existing solutions focus on:
\begin{itemize}[leftmargin=*]
  \item \textbf{Individual prediction analysis}: identifying the most important features of an individual input to explain a model prediction \cite{integrated-gradient, chattopadhyay2019neural, kapishnikov2019xrai, simonyan2013deep, grad-cam, grad-cam++};
  \item \textbf{Training data slicing}: identifying the most influential training samples to impact the model \cite{sagadeeva2021sliceline, bhatt2021divine, koh2017understanding};
  \item \textbf{Model abstraction}: abstracting a simplified model (e.g., SVMs and decision tree) to explain the deep learning model \cite{lime, frosst2017distilling, zhang2019interpreting}.
\end{itemize}

Despite those techniques are useful for explaining a trained model,
few work are proposed to \textit{explain how the model predictions are formed during the training process}.
While the progressive training information can be useful,
it is difficult to abstract the underlying model evolving semantics. 
The semantic questions can be (but not limited to):
(1) how the (re)training process gradually improves the model robustness, and reshapes the classification boundary?
(2) how the model gradually makes a trade-off to fit some samples while sacrificing the others? 
(3) how the model struggles to fit and learn the hard samples?

In this work, we design a time-travelling visualization solution \fullnametool (\tool),
focusing on manifesting the spatio-temporal causality of the training progress of deep learning classifiers.
\tool projects the learned input representation and their classification landscape into a visible low dimensional space,
showing how model predictions are formed during training stages, from both spatial and temporal perspective.
Spatially, \tool visualizes
(1) the layout of learned input representation and
(2) the classification landscape describing the ``territory'' of each class.
Temporally, \tool visualizes
(1) how the classification landscape and the training input representation evolve over the training epochs and
(2) how the new sampled training inputs can reshape the classification boundary.
The spatio-temporal information allows us to
observe training anomalies (e.g. noisy dataset) and
verify some specific training strategies (e.g. effectiveness of active learning sampling strategies).

Comparing to designing measurement to analyze specific sample or model properties (e.g., Shapley value \cite{ancona2019explaining} and hard sample detection \cite{wu2017sampling}),
we design \tool to support \textit{open-ended} exploration.
That is, \tool faithfully reflects how deep models are learned through the training process,
which not only confirm known model properties,
but also support the discovery of unknown phenomena and model defects.



Our approach takes inputs as classifiers trained under different training stages and its training/testing dataset,
then learns \textit{visualization models} (i.e., via an autoencoder) to
(1) project high-dimensional samples into a visible low-dimensional space,
(2) inverse-project low-dimensional points back to high dimensional space (for visualizing classification landscape), and 
(3) ensure that the visualization models can satisfy a set of spatial and temporal constraints. 
We propose four \textit{visualization properties} for any time-travelling visualization solutions,
to preserve
(1) the topological structure between high and low dimensional manifolds,
(2) the distance between training sample representations and latent decision boundary,
(3) the semantics of samples after projection and inverse-projection to low/high-dimensional space, and
(4) the continuity of visualized landscape across the trained classifier in chronological orders.
In summary, we make the following contributions:
\begin{itemize}
  \item We propose time-travelling visualization solution, \fullnametool (or \tool),
  which aims to visualize the classification landscape with spatio-temporal causality,
  to facilitate verifying the model properties and discovering new model behaviors.
  \item We propose four spatial and temporal properties for any time-travelling visualization techniques, 
  and design a deep learning solution to satisfy them,
  for reflecting the classification landscape.
  \item We build our visualization framework \tool to support visualizing various deep classifiers.
  \item We conduct extensive experiments and case studies, showing (1) the effectiveness of \tool to satisfy the properties and (2) how \tool can help understand the training process and diagnose model behaviors.
\end{itemize}
More details of our tool/experiments are at~\cite{dvi}.

\section{Motivating Example}

\begin{figure}[t]
\centering
\begin{subfigure}[b]{0.14\textwidth}
 \includegraphics[width=\textwidth]{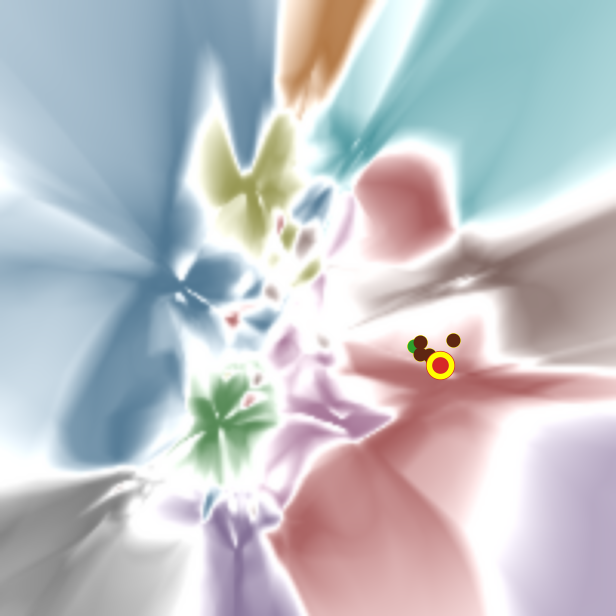}
 \caption{Iteration 1 \\ - adv acc 51.3\% \\ - testing acc 92.3\%}\label{fig:example-adversarial1}
\end{subfigure}
~
\begin{subfigure}[b]{0.14\textwidth}
 \includegraphics[width=\textwidth]{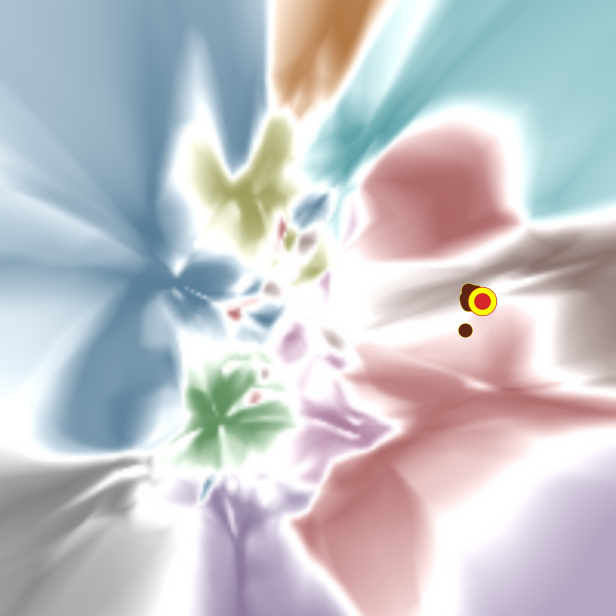}
 \caption{Iteration 2 \\ - adv acc 67.8\% \\ - testing acc 90.3\%}\label{fig:example-adversarial2}
\end{subfigure}
~
\begin{subfigure}[b]{0.14\textwidth}
 \includegraphics[width=\textwidth]{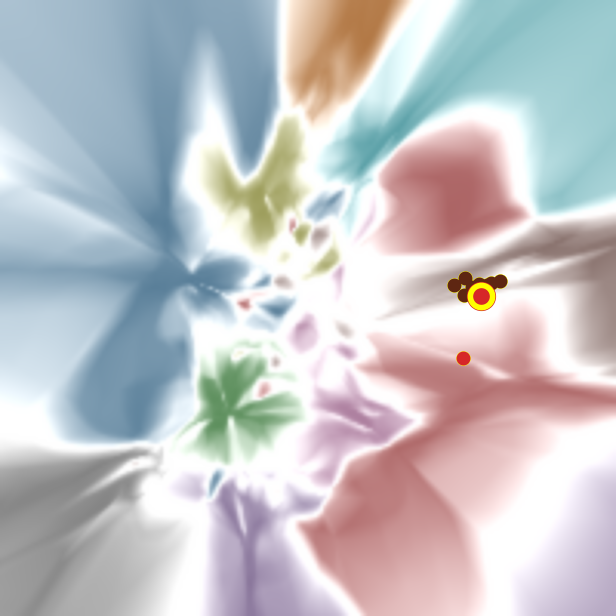}
 \caption{Iteration 3 \\ - adv acc 68.8\% \\ - testing acc 89.9\%}\label{fig:example-adversarial3}
\end{subfigure}
\vspace{-5pt}
\caption{Adversarial training process: dynamics of one testing point and its ten neighbouring adversarial points (adv acc stands for adversarial accuracy, test acc stands for testing accuracy)}\label{fig:adversarial-training-example}
\vspace{-10pt}
\end{figure}

Figure~\ref{fig:adversarial-training-example} shows our visualization of an adversarial training process on CIFAR-10 dataset.
Each point represents a sample and each color represents a class.
The colors of points represent the labels of samples, and the color of a region represent a predicted class.
For example, a point in red (class cat) located in brown (class dog) territory indicates that
it is labelled as cat but classified as dog.
Moreover, the color shade indicates the confidence of prediction,
unconfident regions (i.e. classification boundaries) are visualized as white regions.
Overall, the classification region and boundaries form the \textit{classification landscape}.
Here, \ul{\textit{the model fitting process is visualized by the process of 
(1) classification boundary being reshaped and 
(2) those data points being pulled towards the territory of the corresponding colors}}.

Figure~\ref{fig:adversarial-training-example} shows that \tool manifests
(1) the boundary reshaping process when the model is adapting new adversarial and training samples, and
(2) the process of trade-off being made between adversarial robustness and testing accuracy.
For clarity, we show one testing point (large red point with yellow edge) and its ten nearest neighbour adversarial points (in brown) in Figure~\ref{fig:adversarial-training-example}.
During adversarial training,
(1) the adversarial points are gradually pulled to their color-aligned territory, while
(2) the testing point is also gradually ``pulled'' away from its color-aligned territory to the territory of its adversarial neighbours.
Such trade-off is formed gradually.
In \cite{dvi}, we can further show such trade-off exists by visualizing the dynamics of overall data points.
\tool tool can further visualize the process as animation.
In addition, it supports samples and iteration queries for users to observe the dynamics of interested samples and iterations,
gaining deep insights into the model training process.


\section{Related Work}\label{sec:related-work}
\subsubsection{Explainable AI (XAI) via Attribution Techniques}
To track the causality of (in)correct model predictions,
researchers have proposed approaches to track the prediction back to input, i.e. attribution method~\cite{grad-cam, integrated-gradient, chattopadhyay2019neural, simonyan2013deep, shrikumar2016not, kapishnikov2019xrai}.
Attribution solutions evaluate the contribution of any input components (e.g. some pixels in the image) to the prediction outcome.
\cite{integrated-gradient} proposed two axioms that every attribution method should satisfy, and developed integrated gradients(IG).
Chattopadhyay et al. proposed average causal effect (ACE)~\cite{chattopadhyay2019neural} to mitigate the bias introduced by IG.
To visualize the attribution explanation,
Selvaraju et al. proposes the Grad-Cam~\cite{grad-cam} solution to highlight the pixels on an input image to explain the prediction.

Different from those approaches explaining an individual sample,
\tool visualizes the process how the classification landscape is formed.
\tool and attribution analysis are complementary.
Users can use \tool to observe an overview of classification landscape and the distribution of the input samples,
then use any attribution technique to inspect individual samples.


\subsubsection{Model Visualization}
Typically, model visualization is transformed to a dimension reduction problem. 
Existing techniques include linear methods (e.g. PCA~\cite{pca}, LDA~\cite{lda}, etc) and non-linear methods (e.g. t-SNE~\cite{t-sne}, UMAP~\cite{umap}.
Non-linear solutions preserve the neighbor relations after projecting data to a low-dimensional space.
To this end,
Van der Maaten at al. proposed t-SNE, which transforms the distance of high-dimensional samples into a conditional probability with Gaussian distribution
and that of low-dimensional samples into a conditional probability with Student t-distribution~\cite{t-sne} as similarity measurements.
Tang et al. and McInnes at al. propose LargeViz~\cite{largeviz} and UMAP~\cite{umap} to further improves the performance.
Different from \tool, they visualize sample layout instead of the classification landscape.

One relevant work is DeepView~\cite{deepview}, aiming to visualize the decision boundaries of a classifier.
DeepView projects high-dimensional sample into low-dimensional space via UMAP,
with a customized manifold distance regarding the prediction outcome and the Euclidean distance in the input space.
DeepView inverse-projects a low-dimensional point regarding the high-dimensional counterparts of its neighbours.
\tool is different from DeepView in two folds.
First, \tool is way more efficient and scalable than DeepView (see Section~\ref{sec:experiment}).
Second, \tool considers boundary-preserving property and temporal property, which are essential in time-travelling visualization.


\section{Properties of Time-Travelling Visualization}\label{sec:properties}

\begin{table}[t]
\centering
\small
\caption{Notation table for $C$-class classification task}
\vspace{-5pt}
\label{tab:notation}
\begin{tabular}{p{1cm}p{3.5cm}p{2.5cm}}
\toprule
Notation & Definition & Dimension \\
\hline
$\mathbf{S/X/Y}$  & Training data inputs,  representations, and low-dimensional embeddings & $\mathbb{R}^{N \times d}, \mathbb{R}^{N \times h}, \mathbb{R}^{N \times l}$  \\
\hline
$\mathcal{S/X/Y}$  & Input space,  manifold space, low-dimensional embedding space &  $\mathcal{S} \subset \mathbb{R}^d,  \mathcal{X} \subset \mathbb{R}^h,  \mathcal{Y} = \mathbb{R}^l$\\
\hline
$\phi(.)$  & Projection function  & $\mathbb{R}^h \rightarrow \mathbb{R}^l$  \\
\hline
$\psi(.)$ & Inverse-projection function & $\mathbb{R}^l \rightarrow \mathbb{R}^h$ \\
\hline
$f(.)$ & Feature function & $\mathbb{R}^{d} \rightarrow \mathbb{R}^h$ \\
\hline
$g(.)$ & Prediction function & $\mathbb{R}^h \rightarrow \mathbb{R}^C$                                            \\
\hline
$c(.)$ & Classifier, i.e. $g(f(.))$ & $\mathbb{R}^{d}\rightarrow\mathbb{R}^C$ \\
\hline
$\mathbf{B}$ & Boundary points in $\mathbb{R}^h$ & $\forall \mathbf{b}_i\in \mathbf B, \mathbf{b}_i\in \mathbb{R}^h$ \\
\bottomrule
\end{tabular}
\vspace{-10pt}
\end{table}

In this section, we propose four properties for any time-travelling visualization techniques. 


\subsection{Notation Definition}
We use the notation in Table~\ref{tab:notation}.
We have a subject model $c(.)$ for a $C$-class classification problem.
The input space is denoted as $\mathcal{S}$
where $\mathcal{S} \subset \mathbb{R}^d$.
$\mathbf{S} = [\mathbf{s}_1, \mathbf{s}_2, ... \mathbf{s}_N]^T$ is training input set.
$f:\mathbb{R}^d \rightarrow\mathbb{R}^h$ is a feature function, such that $\mathbf{x}=f(\mathbf{s})$ is a representation vector with $h$ dimensions
for an input $\mathbf{s}\in\mathbf{S}$.
We denote the manifold space of the representation vectors as $\mathcal{X}$ where $\mathcal{X} \subset \mathbb{R}^h$.
The learnt representations for training data is denoted as $\mathbf{X}$ where $\mathbf{X} = [\mathbf{x}_1, \mathbf{x}_2, ... \mathbf{x}_N]^T$.
Let $g:\mathbb{R}^h\rightarrow\mathbb{R}^C$ be the prediction function, where $g(\mathbf{x})_i$ represents the logits for $i^{th}$ class.
A classifier $c$ consists of $f$ and $g$, i.e. $c = g \circ f:\mathbb{R}^d \rightarrow \mathbb{R}^C$.
Taking $c$ and its training inputs,
we derive a visualization model $V=\langle \phi, \psi \rangle$:
\begin{itemize}[leftmargin=*]
  \item A projection function $\phi:\mathbb{R}^h \rightarrow \mathbb{R}^l$, which projects manifold space $\mathcal{X}$ to a visible low-dimensional space $\mathcal{Y}$ where $\mathcal{Y} = \mathbb{R}^l$ ($l$ is 2 or 3).
  Projecting $\mathbf{X}$ on to $\mathcal{Y}$ (i.e. $\mathbf{Y} = \phi(\mathbf{X})$) produces their counterparts $\mathbf{Y} = [\mathbf{y}_1, \mathbf{y}_2, ... \mathbf{y}_N]^T$.
  \item An inverse-projection function $\psi: \mathbb{R}^l \rightarrow \mathbb{R}^h$, which inverse-projects visible low-dimensional space $\mathcal{Y}$ back to representation space $\mathcal{X}$.
\end{itemize}

\subsection{Neighbour Preserving Property}
\begin{definition}[k-witness]\label{def:k-witness}
Given a training dataset $\mathbf{S}$ and a distance metric defined on $\mathbf{X}$,
$d :\mathbb{R}^h\times\mathbb{R}^h \rightarrow \mathbb{R}_{\geq 0}$.
For a given $\mathbf{x}_i\in\mathbf{X}$, we denote the \ul{index set} of its k-nearest neighbors as
$N_k(\mathbf{x}_i) = \argmin_{\mathcal{J} \subset \{1..N\}\backslash\{i\}, |\mathcal{J}|=k}{\sum_{j \in \mathcal{J}} d(\mathbf{x}_j, \mathbf{x}_i)}$.
We say $\mathbf{x}_j$ is k-witnessed by $\mathbf{x}_i$ in $\mathbf{X}$ if $j\in N_k(\mathbf{x}_i)$.
\end{definition}

Given a data sample $\mathbf{s}$, with its representation being $\mathbf{x}\in \mathbf{X}$ and low-dimensional counterpart being $\mathbf{y} \in \mathbf{Y}$,
any $\mathbf{x}'$ being k-witnessed by $\mathbf{x}$ should have its counterpart $\mathbf{y}'$ being k-witnessed by $\mathbf{y}$,
and vice versa.

Assuming the manifold $\mathcal{X}$ of $\mathbf{X}$ is known, we denote the distance between $\mathbf{x}_i$ and $\mathbf{x}_j$ in manifold as $d_\mathcal{M} (\mathbf{x}_i, \mathbf{x}_j)$.
Similarly, we denote the distance of their counterparts $\mathbf{y}_i$ and $\mathbf{y}_j$ as $d_\mathcal{E}(\mathbf{y}_i, \mathbf{y}_j)$ in Euclidean space.
Given a witness value $k$, we define
$N_k(\mathbf{x}_i)$ $:=$ $\argmin_{\mathcal{J} \subset \{1..N\}\backslash\{i\}, |\mathcal{J}|=k}{\sum_{j \in \mathcal{J}} d_\mathcal{M}(\mathbf{x}_j, \mathbf{x}_i)}$ and
$N_k(\mathbf{y}_i)$ $:=$ $\argmin_{\mathcal{J} \subset \{1..N\}\backslash\{i\}, |\mathcal{J}|=k}{\sum_{j \in \mathcal{J}} d_\mathcal{E}(\mathbf{y}_j, \mathbf{y}_i)}$,
representing two index sets of neighbours being $k$-witnessed by $\mathbf{x}_i$ and its counterpart $\mathbf{y}_i$ respectively.
The neighbour-preserving property requires to maximize the $k$ spatial neighbour preserving rate
\begin{equation}\label{eq:spatial-neighbour-preserving-rate}
   nn_{pv}(k) := \frac{1}{N}\sum_{i=1}^{N} \frac{|N_k(\mathbf{x}_i) \cap N_k(\mathbf{y}_i)|}{k}
\end{equation}


\subsection{Boundary Distance Preserving Property}\label{sec:boundary-property}

\begin{definition}[$\delta$-Boundary]\label{def:boundary}
For a small $\delta\in[0, 1)$, a prediction function $g:\mathbb{R}^h\rightarrow\mathbb{R}^C$ and a min-max rescaling function $r:\mathbb{R}^C\rightarrow [0, 1]^C$, let $r(g(\mathbf{x}))_{top1}$ and $r(g(\mathbf{x}))_{top2}$ be the largest and second largest value of $r(g(\mathbf{x}))$ respectively.
We say that a point $x$ lies on $\delta$-Boundary if $\left|r(g(\mathbf{x}))_{top1} - r(g(\mathbf{x}))_{top2}\right| \leq \delta$.
\end{definition}
We define classification boundary as a set of points $\mathbf{B}=\{\mathbf{b}|\mathbf{b}$ is on $\delta$-boundary$\}$.
Similar to neighbour preserving property, the boundary distance preserving property requires that any $\mathbf{x}_i\in \mathbf{X}$ should preserve its $k$ nearest boundary neighbours after being projected to $\mathbf{y}_i$ by $\phi(.)$.
If we denote $\mathbf{b}$ as a boundary point in $\mathbb{R}^h$, and its counterpart in $\mathbb{R}^l$ as $\mathbf{b}'$.
Extending Definition \ref{def:k-witness},
we define $N^{(b)}_k(\mathbf{x}_i) := \argmin_{\mathcal{J} \subset \{1..|\mathbf{B}|\}, |\mathcal{J}|=k}{\sum_{j \in \mathcal{J}} d_\mathcal{M}(\mathbf{b}_j, \mathbf{x}_i)}$,
$N^{(b')}_k(\mathbf{y}_i) := \argmin_{\mathcal{J} \subset \{1..|\mathbf{B}|\}, |\mathcal{J}|=k}{\sum_{j \in \mathcal{J}} d_\mathcal{E}(\mathbf{b}'_j, \mathbf{y}_i)}$
representing two index sets of being k-boundary-witnessed by $\mathbf{x}_i$ and its counterpart $\mathbf{y}_i$.
We require the projection function $\phi(.)$ should maximize:
\begin{equation}\label{eq:boundary-distance-preserving-property}
  boundary_{pv}(k) := \frac{1}{nnN}\sum_{i=1}^{N}\frac{|N^{(b)}_k(\mathbf{x}_i) \cap N^{(b')}_k(\mathbf{y}_i)|}{k}
\end{equation}


\subsection{Inverse-Projection Preserving Property}

To visualize the classification landscape,
the visualization solution needs an inverse projection function $\psi(.)$
to reconstruct high-dimensional representation vectors from low-dimensional vectors in $\mathcal{Y}$.
Such a reconstruction needs to satisfy that
(1) any low-dimensional vector $\mathbf{y}_i$ projected from a representation vector $\mathbf{x}_i$, should be reconstructed to a $\mathbf{x}'_i$ as close to $\mathbf{x}_i$ as possible; and
(2) it can generalize to arbitrary low-dimensional vectors.
The first requirement ensures that the projection cause little information loss.
Moreover, when representing each class as a distinct color, the second requirement allows us to \textit{taint} arbitrary points in a low-dimensional canvas.
Given $\mathbf{H}=\{\mathbf{h_i} | \mathbf{h_i} \in \mathcal{X}\}$, this property requires that $\psi(.)$ can minimize the reconstruction error:
\begin{equation}\label{eq:inverse-projection-preserving-property}
  rec_{pv} := \frac{1}{|\mathbf{H}|}\sum_{i=1}^{|\mathbf{H}|} \norm{\mathbf{h}_i - \psi(\phi(\mathbf{h}_i))}^2
\end{equation}

\subsection{Temporal Preserving Property}
Different from existing \textit{static} visualization as UMAP and t-SNE,
our visualized classification landscape requires to preserve the temporal continuity of the classification landscape change of the subject classifier.
Assuming that two classifiers $c^t$ and $c^{t+1}$ are classifiers trained in two consecutive epochs,
their classification landscapes are supposed to be similar.
Thus, their visualization solutions $V^t$ and $V^{t+1}$ should provide similar visualization results.

We consider
(1) classifiers $c^t=g^t \circ f^t$ and $c^{t+1}=g^{t+1} \circ f^{t+1}$ taken in chronological order, and
(2) a measurement function $eval_{sem}(\cdot)$ to evaluate the semantic similarity of an input $\mathbf{s}\in\mathbf{S}$ in the representation space $\mathcal{X}$.
We define the input semantic as the number of its shared $k$-witnessed neighbors between consecutive epochs.
Let $N_k(\mathbf{x}^t_{i})$ be the index set of all points being k-witnessed by $\mathbf{x}_i$ in step $t$, and $N_k(\mathbf{x}^{t+1}_{i})$ in step $t+1$, $eval_{sem}$ is defined as:
\begin{equation}
    \begin{split}\label{eq:eval_sem}
    & eval_{sem}(\mathbf{x}^{t}_{i}, \mathbf{x}^{t+1}_{i}, k) := \frac{|N_k(\mathbf{x}^{t}_{i}) \cap N_k(\mathbf{x}^{t+1}_{i})|}{k}
    \end{split}
\end{equation}
If two epochs have similar semantics,
the visualization solutions $V^{t}$ and $V^{t+1}$ should project $\mathbf{x}^{t}$ and $\mathbf{x}^{t+1}$ to similar positions in $\mathbb{R}^l$,
or have a negative correlation with $d_\mathcal{E}(\phi^t(\mathbf{x}^{t}), \phi^{t+1}(\mathbf{x}^{t+1}))$.
We define the correlation as:
\begin{equation}
    \begin{split}\label{eq:corr}
    & temporal_{pv}(k) := \\ & corr(eval_{sem}(\mathbf{x}^{t}, \mathbf{x}^{t+1}, k), d_\mathcal{E}(\phi_{t}(\mathbf{x}^{t}), \phi^{t+1}(\mathbf{x}^{t+1})))\\
    \end{split}
\end{equation}
Then we require projection function $\phi^{t}(\cdot)$ and $\phi^{t+1}(\cdot)$ to minimize $temporal_{pv}(k)$:

To the best of our knowledge, none of the existing approaches have addressed all four properties.
t-SNE and UMAP only satisfy the neighbour preserving property;
DeepView satisfies the neighbour preserving and the inverse-preserving property.
We make the first solution regarding all four properties.

\section{Approach}\label{sec:approach}
\subsubsection{Overview}
\begin{figure}
  \centering
  \includegraphics[scale=0.6]{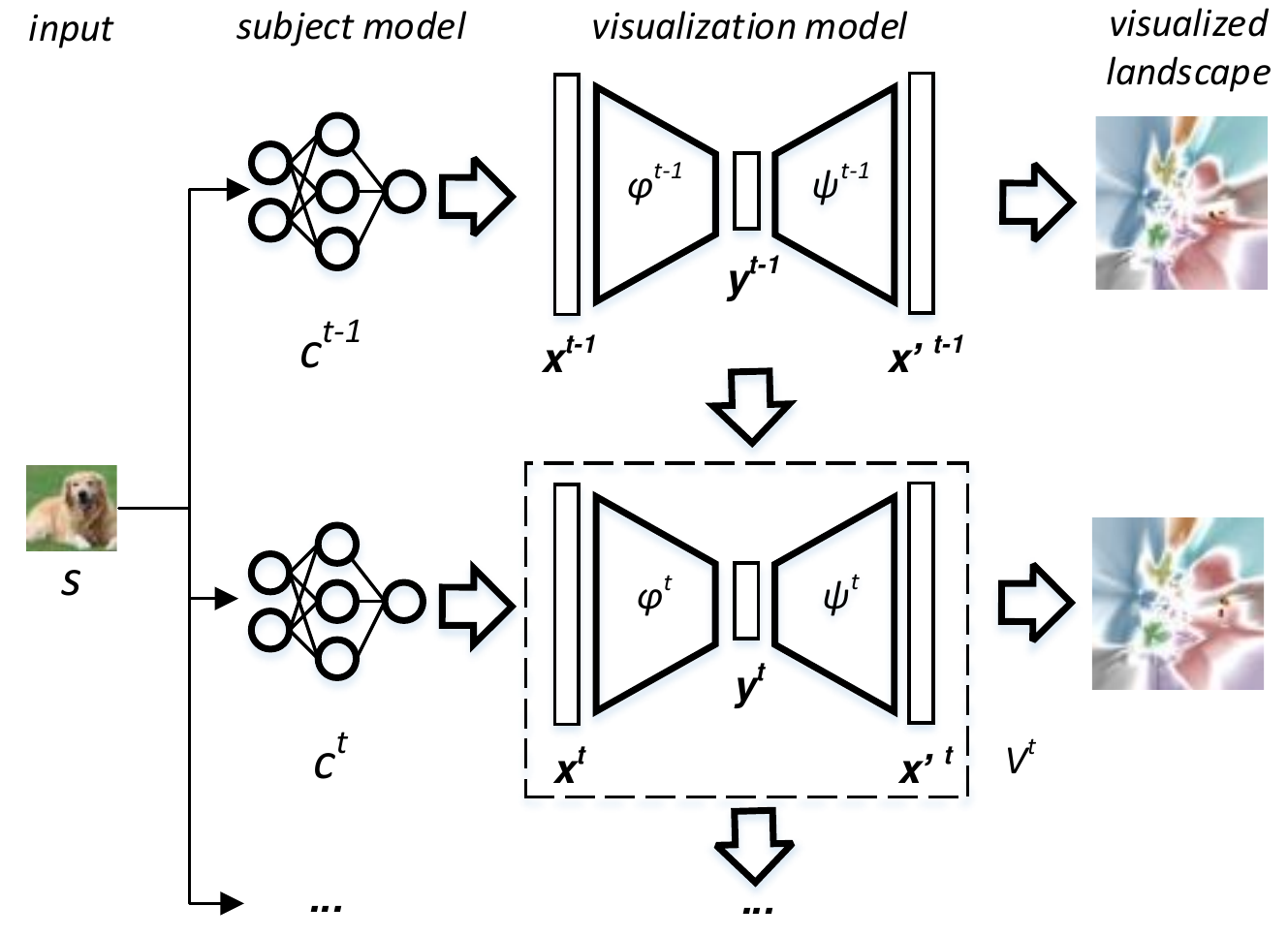}
  \caption{Overview of \fullnametool}\label{fig:overview}
  \vspace{-15pt}
\end{figure}

As showed in Figure~\ref{fig:overview}, \tool
takes as input a sequence of classifiers trained in chronological order, $\mathbf{C}=\left\{c^1, c^2, ..., c^T\right\}$ as subject models, and
generates a corresponding sequence of visualization models (i.e. autoencoders) $\mathbf{V}=\left\{V^1, V^2, ..., V^T\right\}$ to derive visualized classification landscape.
We use superscript to denote the chronological order of all notations.
For each visualization model $V^t = \langle \phi^t, \psi^t \rangle$, the encoder serves as projection function $\phi^t$ and decoder as inverse-projection function
$\psi^t$.

Each class is given with a non-white color,
$V^t$ can
(1) calculate the coordinate of each input $\mathbf{s}\in\mathbf{S}$ via $\phi^t(f^t(\mathbf{s}))$, and
(2) taint arbitrary point $\mathbf{y}$ via $g^t(\psi^t(\mathbf{y}))$.
If $\mathbf{y}$ lies on $\delta$-boundary (see Definition~\ref{def:boundary}), it is tainted in white; otherwise, it is tainted in the representing color of class $g^t(\psi^t(\mathbf{y}))_{top1}$.

Each visualization model $V^{t}$ for $c^t$ is trained regarding the four spatial and temporal properties.
We
(1) estimate representative $\delta$-boundary points for $c^t$;
(2) construct a topological complex for boundary/training representation vectors and
preserve its structure after projection to satisfy (boundary) neighbour-preserving property;
(3) minimize the distance between $\mathbf{x}$ and reconstructed $\psi^t(\phi^t(\mathbf{x}))$ to satisfy the inverse-projection preserving property; and
(4) build the continuity between (1) $\phi^t$ and $\phi^{t-1}$ and (2) $\psi^t$ and $\psi^{t-1}$ ($t \ge 2$) to satisfy the temporal-preserving property.

\subsection{$\delta$-Boundary Estimation.}
We estimate $\delta$-boundary by synthesizing boundary samples,
regarding the efficiency, authenticity, and diversity.

\subsubsection{Efficiency and Authenticity}
\begin{figure}[t]
\centering
\begin{subfigure}[b]{0.13\textwidth}
 \includegraphics[width=\textwidth]{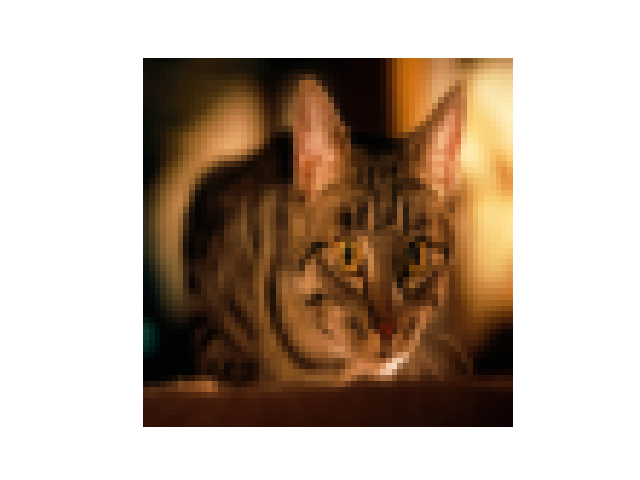}
 \caption{Image 1}\label{fig:example-adversarial1}
\end{subfigure}
~
\begin{subfigure}[b]{0.13\textwidth}
 \includegraphics[width=\textwidth]{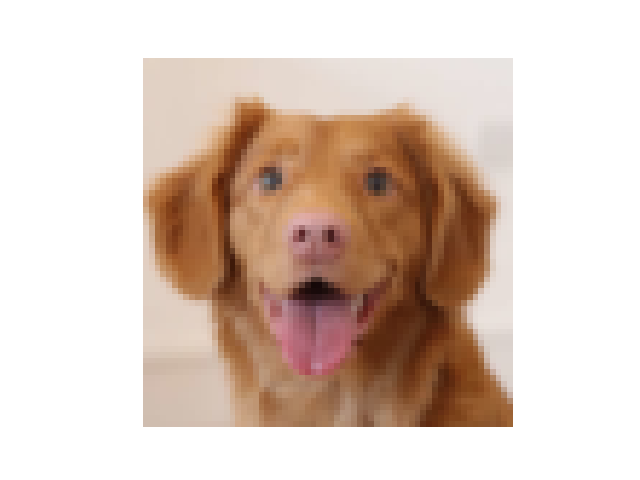}
 \caption{Image 2}\label{fig:example-adversarial2}
\end{subfigure}
~
\begin{subfigure}[b]{0.13\textwidth}
 \includegraphics[width=\textwidth]{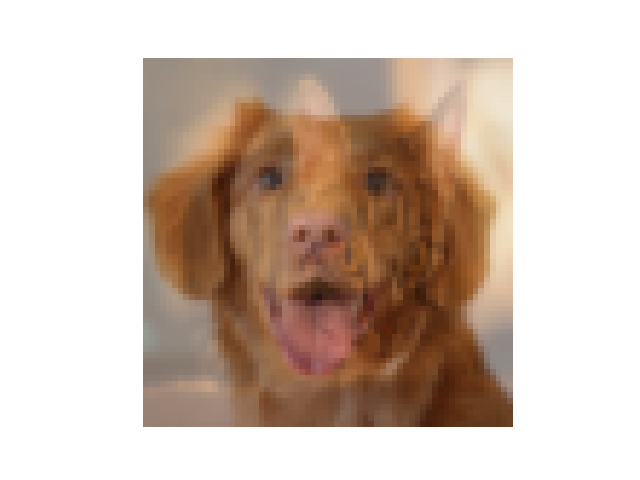}
 \caption{Mixed-up}\label{fig:example-adversarial3}
\end{subfigure}
\vspace{-5pt}
\caption{Example of mixed-up image, $\lambda$=0.35}\label{fig:mixed-up}
\vspace{-15pt}
\end{figure}

We propose a novel mixup-based point synthesis method.
Given a classifier $c(.)$ and two input images $\mathbf{s}_i, \mathbf{s}_j \in \mathbf{S}$ from two different predicted classes,
our rationale lies in that,
\begin{enumerate}[leftmargin=*]
  \item Their mixed-up inputs (i.e. images) can still largely preserve its inherent distribution of $\mathcal{S}$ (see Figure~\ref{fig:mixed-up});
  \item Assuming continuity of $c(.)$, and
    the linear interpolation $\mathbf{s}_b = \lambda\cdot \mathbf{s}_i + (1-\lambda)\cdot \mathbf{s}_j, \lambda \in [0,1]$,
    we can find a $\lambda$ such that $\mathbf{s}_b$ lies on the $\delta$-boundary within $\mathcal{O}(\log_2 \frac{d_{\mathcal{E}}(\mathbf{s}_i, \mathbf{s}_j)}{width(\delta)})$ rounds of binary search,
    where $width(\delta)$ is the width of $\delta$-boundary on the line segment connecting $\mathbf{s}_i$ and $\mathbf{s}_j$ in Euclidean space.
\end{enumerate}

To synthesize an authentic boundary sample, we set an upper bound for $\lambda$.
Comparing to adversarial sample generation techniques (e.g., Deepfool~\cite{moosavidezfooli2016deepfool}) which require expensive search overhead and highly depend on the model gradients,
our mixup-based approach has more guarantee to synthesize a boundary sample within a limited search budget.

\subsubsection{Diversity}
Given $C$ classes in a classifier $c(.)$, we synthesize boundary samples for $C \choose 2$ pairs of classes.
Regarding both diversity and efficiency of synthesis,
we favour the pairs (1) with less number of boundary samples generated so far and (2) with high successful synthesis rate.
Specifically,
\begin{equation}\label{eq:samplepairprob}
    \begin{split}
    \Pr(p=(C_i, C_j)) & = \alpha \cdot \Pr(s(C_i, C_j)) \\
     & + (1-\alpha) \cdot succ(C_i, C_j)
    \end{split}
\end{equation}

In Equation~\ref{eq:samplepairprob}, we introduce a trade-off parameter $\alpha\in[0, 1]$, between $\Pr(s(C_i, C_j))$
(i.e., the relative boundary sample abundance) and
$succ(C_i, C_j)$ (i.e., the success rate to synthesize a boundary point).
Specifically,
\begin{equation}\label{eq:diversity-sampling}
    \Pr(s(C_i,C_j)) = \frac{max(0, \rho-num(C_i,C_j))}{\sum_{k\neq m}max(0,(\rho-num(C_k, C_m)))}
\end{equation}
$num((C_i, C_j))$ is the generated boundary points between class $C_i$ and $C_j$ so far, and $\rho$ is the mean number of generated point over all pairs of classes.

We estimate the successful synthesis rate of a pair as:
\begin{equation}\label{eq:efficiency-sampling}
    succ(C_i, C_j) = \frac{num_{b}(C_i, C_j)}{num_{syn}(C_i, C_j)}
\end{equation}
$num_{syn}(\cdot)$ is the number of trials to synthesize boundary between a pair and $num_{b}(\cdot)$ is the number of successful trials within a search budget.

\subsection{(k)-BAVR Complex construction}
Given representation vector set $\mathbf{X}$ and its derived boundary vectors $\mathbf{B}$,
we construct a (k)-Boundary-Augmented Vietoris-Rips complex on $\mathbf{U} := \mathbf{X} \cup \mathbf{B}$,
to sample a representative subset of edges $(p_i, p_j)\in \mathbf{U} \times \mathbf{U}$ for training an encoder to preserve boundary/non-boundary neighbors.

\begin{definition}[(k)-Boundary-Augmented-Vietoris-Rips Complex]\label{def:complex}
A (k)-Boundary-Augmented-Vietoris-Rips Complex ((k)-BAVR Complex) ($k > 0$) is a simplicial complex consisting of 0-simplices and 1-simplices such that
(1) each 0-simplex is a point from $\mathbf{U} := \mathbf{X} \cup \mathbf{B}$, and
(2) each 1-simplex consists of two points in $\mathbf{U}$ and their connecting edge, satisfying one of the following conditions:
\begin{enumerate}[leftmargin=*]
  \item[(a)] $\left\{(\mathbf{x}_i,\mathbf{x}_j):\forall \mathbf{x}_i\in\mathbf{X}, j\in N_k(\mathbf{x}_i)\right\}$ where $N_k(\mathbf{x}_i)$ is the index set of points that are k-witnessed by $\mathbf{x}_i$ in $\mathbf{X}$.
  \item[(b)] $\left\{(\mathbf{x}_i,\mathbf{b}_j):\forall \mathbf{x}_i\in\mathbf{X}, j \in N^{(b)}_k(\mathbf{x}_i)\right\}$, where $N^{(b)}_k(\mathbf{x}_i)$ is the index set of points being k-boundary-witnessed by $\mathbf{x}_i$.
  \item[(c)] $\left\{(\mathbf{b}_i,\mathbf{b}_j):\forall \mathbf{b}_i\in\mathbf{B}, j \in N_k(\mathbf{b}_i)\right\}$, where $N_k(\mathbf{b}_i)$ is the index set of $\mathbf{b}_i$'s k nearest boundary neighbors.
\end{enumerate}
\end{definition}

Intuitively, (k)-BAVR Complex captures the topological structure of $\mathbf{U} := \mathbf{X} \cup \mathbf{B}$.
Based on the complex, we sample a positive pair set $P_{x \times x +} \subset \mathbf{X} \times \mathbf{X}$
where $p=(\mathbf{x}_i, \mathbf{x}_j)\in P_{x\times x +}$ so that $\mathbf{x}_i$ and $\mathbf{x}_j$ form a 1-simplex.
Similarly, we obtain $P_{x\times b +} \subset \mathbf{X}\times\mathbf{B}$ and $P_{b\times b +} \subset \mathbf{B}\times\mathbf{B}$.
In addition, we randomly choose pairs from $\mathbf{X}\times\mathbf{X}$, $\mathbf{X}\times\mathbf{B}$, and $\mathbf{B}\times\mathbf{B}$ to construct three negative pair sets, i.e., $P_{x\times x -} \subset \mathbf{X}\times\mathbf{X}$, $P_{x\times b -} \subset \mathbf{X}\times\mathbf{B}$, and $P_{b\times b -} \subset \mathbf{B}\times\mathbf{B}$.

Finally, given $P = P_{x\times x +} \cup P_{x\times x -} \cup P_{x\times b +} \cup P_{x\times b -} \cup P_{b\times b +} \cup P_{b\times b -}$,
we follow the parametric umap loss function defined in \cite{umap} and \cite{sainburg2020parametric} to train our encoder $\phi$.

\subsection{Inverse-Projection Preserving}
We design our loss function to train the encoder $\phi$ and the decoder $\psi$ as:
\begin{equation}\label{eq:reconstruction-loss}
  \mathcal{L}_{rec} := \frac{1}{Nh}\sum_{i=1}^{N}\sum_{m=1}^{h}(1+grad_i^m)^{\beta} ||\mathbf{x}_i^m - \psi(\phi(\mathbf{x}_i^m))||^2
\end{equation}
\begin{equation}
    grad_i := abs(\frac{\partial g(\mathbf{x}_i)_{top1}}{\partial \mathbf{x}_i})+
    abs(\frac{\partial g(\mathbf{x}_i)_{top2}}{\partial \mathbf{x}_i})
\end{equation}
where
$h$ is the number of dimensions,
$g(\mathbf{x}_i)_{top1}$ is the largest value in $g(\mathbf{x}_i)$ and $g(\mathbf{x}_i)_{top2}$ is the second largest value in $g(\mathbf{x}_i)$.
The rationale lies in that
we need to preserve the most critical information of representation vector $\mathbf{x}$ after projecting and inverse-projecting back to the original space.
In this work, such information lies in the top-1 dimension of $g(.)$ (for predicting its class) and $g(\mathbf{x}_i)_{top2}$ (for measuring the boundary).
By tracking the gradients from $g(\mathbf{x}_i)_{top1}$ and $g(\mathbf{x}_i)_{top2}$,
we can force the encoder and decoder to learn such information.

\subsection{Temporal Continuity}
We preserve the temporal continuity with transfer learning and a temporal loss function.
Given $V^{t-1} (t > 1)$, $V^{t}$ is initialized with $V^{t-1}$'s weights.
We bound the change of $V^{t}$ from $V^{t-1}$ by defining a temporal loss regarding the \textit{temporal neighbour preserving rate}.
\begin{equation}\label{eq:temporal-loss}
  \mathcal{L}_{t} := \frac{1}{N}\sum_{i=1}^N eval_{sem}(\mathbf{x}^{t-1}_{i}, \mathbf{x}^{t}_{i}, k) \cdot \norm{\mathbf{W}_{t} - \mathbf{W}_{t-1}}^2
\end{equation}

In Equation~\ref{eq:temporal-loss}, $\mathbf{W}_{t}$ is the weights of the $\phi(.)$ and $\psi(.)$, while $\mathbf{W}_{t-1}$ is the weights of $\phi(.)$ and $\psi(.)$ learned in previous epoch.
The final loss function to train $\phi(.)$ and $\psi(.)$ is the weighted sum of all the loss functions, i.e.,
\begin{equation}\label{eq:loss}
  \mathcal{L}_{total} = \lambda_1 \cdot \mathcal{L}_{umap} + \lambda_2 \cdot \mathcal{L}_{rec} + \lambda_3 \cdot \mathbb{1}(t>1) \cdot \mathcal{L}_{t}
\end{equation} 
\section{Evaluation}\label{sec:experiment}

\noindent\textbf{Property Measurement.}
We measure the spatial and temporal properties, i.e.,
$nn_{pv}(k)$, $boundary_{pv}(k)$, $rec_{pv}$, and $temporal_{pv}(k)$ (see Section~\ref{sec:properties}) as follows.
\begin{itemize}[leftmargin=*]
  \item \textbf{Preserving Neighbour and Boundary Distance}:
    We use $nn_{pv}(k)$ and $boundary_{pv}(k)$, and let $k=10, 15, 20$. 
  \item \textbf{Preserving Inverse-Projection}:
    We evaluate the prediction preserving rate, i.e., $PPR := \frac{|\mathcal{Q}|}{N}$ where
    $\mathcal{Q} := \{\mathbf{x} | \argmax_c{g_c(\mathbf{x})} = \argmax_c{g_c(\psi(\phi(\mathbf{x})))}, \mathbf{x}\in \mathbf{X}\}$.
  \item \textbf{Preserving Temporal Continuity}:
    For $temporal_{pv}(k)$, we use Pearson correlation and set $k$ as 10, 15, and 20.
\end{itemize}

\noindent\textbf{Dataset and Subject Model.}
We choose three datasets, i.e., MNIST (60K/10K training/testing set), Fashion-MNIST (60K/10K training/testing set), and CIFAR-10 (50K/10K training/testing set).
We use ResNet18~\cite{he2016deep} as the subject classifier,
and global average pooling layer as the feature vector (i.e., 512 dimensions).

\noindent\textbf{Baseline.}
We select PCA, t-SNE, UMAP, and DeepView as baselines.
We compare \tool with PCA, t-SNE, UMAP on the whole datasets.
The implementation of DeepView has limitation on its scalability over 1000 samples,
we compare \tool with DeepView by training on 1000 samples (a empirical size suitable for DeepView).
Different from PCA, t-SNE, and UMAP, we repeat the experiment with DeepView for 10 times to mitigate the bias.
Moreover, we randomly select 200 samples as the test set shared by multiple trials.

\noindent\textbf{Runtime Configuration.}
We design our autoencoder as follows.
Given the dimension of the feature vector is $h$,
we let the encoder and decoder to have shape $(h, \frac{h}{2}, \frac{h}{2}, \frac{h}{2}, \frac{h}{2}, 2)$;
and $(2, \frac{h}{2}, \frac{h}{2}, \frac{h}{2}, \frac{h}{2}, h)$ respectively.
Learning rate is initialized with 0.01 and decay every 8 epochs by factor of 10.
The threshold $\delta$ to decide boundary point is set to be 0.1.
We generate $0.1*N$ boundary points,
shared by all the solutions. The upper bound for $\lambda$ in boundary point generation is set to 0.4, $\alpha$ in Equation~\ref{eq:samplepairprob} to 0.8, $\beta$ in Equation~\ref{eq:reconstruction-loss} to 1.0, and the trade-off hyper-parameters in total loss (Equation~\ref{eq:loss}) to 1.0, 1.0, 0.3 respectively.

\begin{figure}[t]
    \centering
    \includegraphics[width=0.47\textwidth]{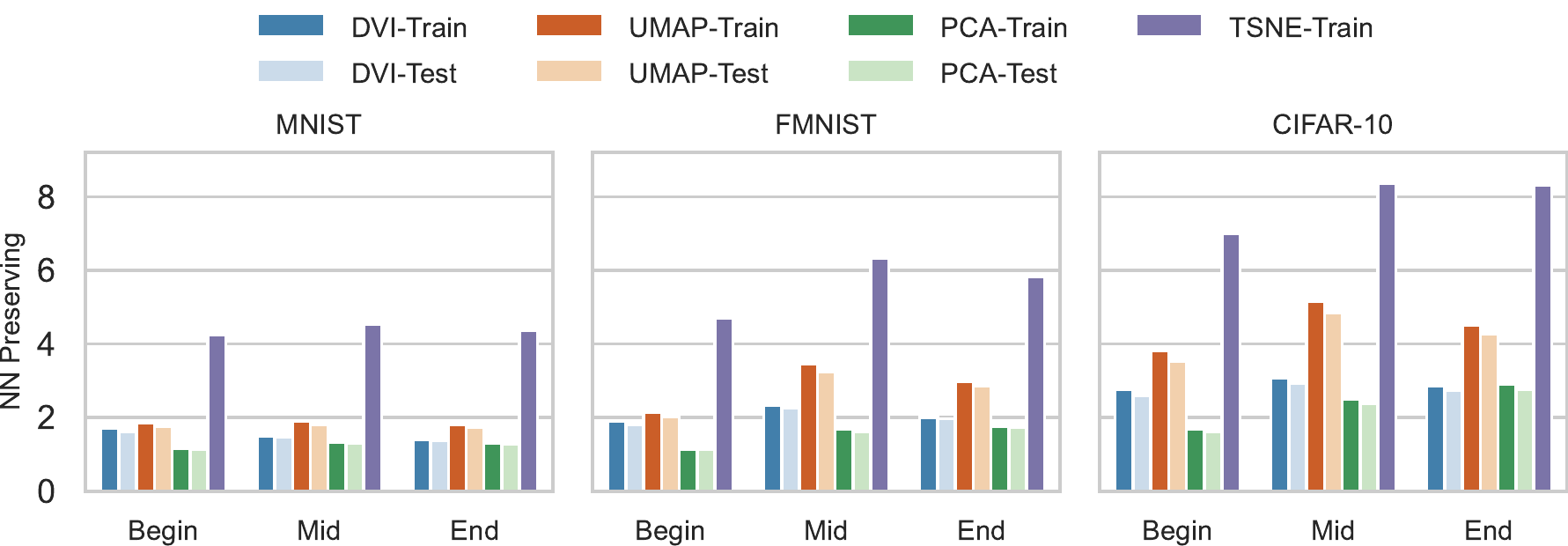}
    \caption{$k$ Neighbour Preserving ($k$=15)}
    \vspace{-5pt}
    \label{fig:nn}
\end{figure}

\begin{figure}[t]
    \centering
    \includegraphics[width=0.47\textwidth]{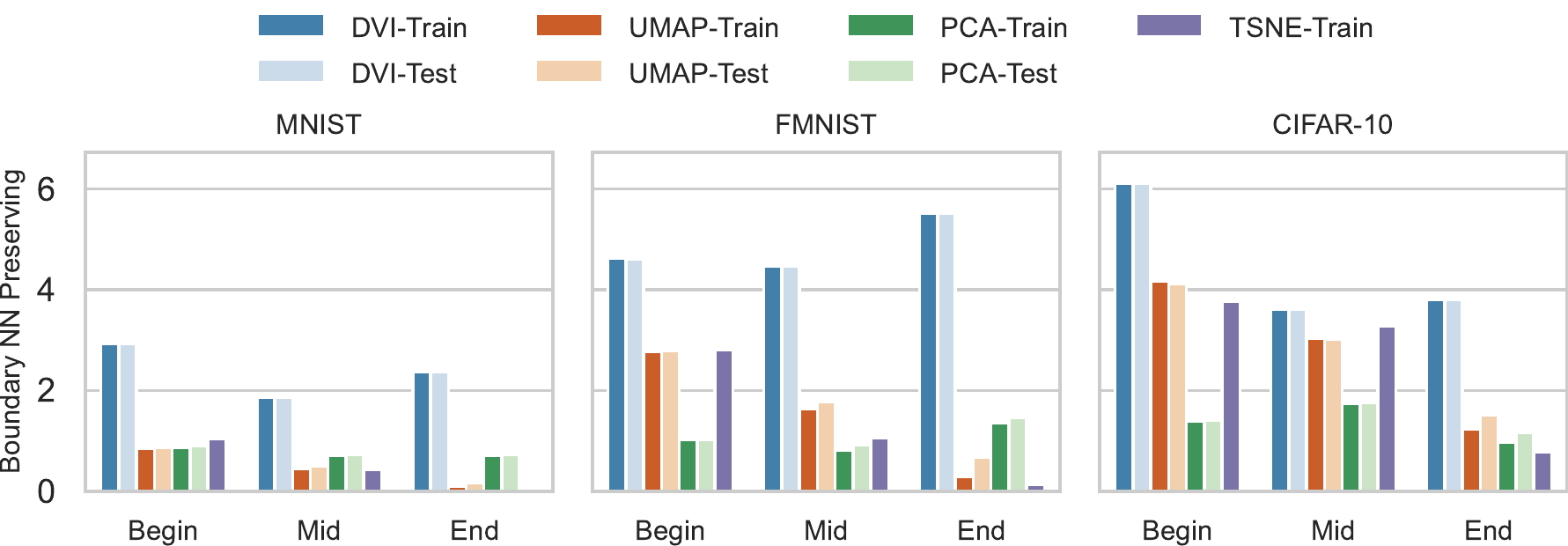}
    \caption{$k$ Boundary Neighbour Preserving ($k$=15)}
    \label{fig:boundary}
    \vspace{-5pt}
\end{figure}
\begin{figure}[t]
    \centering
    \includegraphics[width=0.47\textwidth]{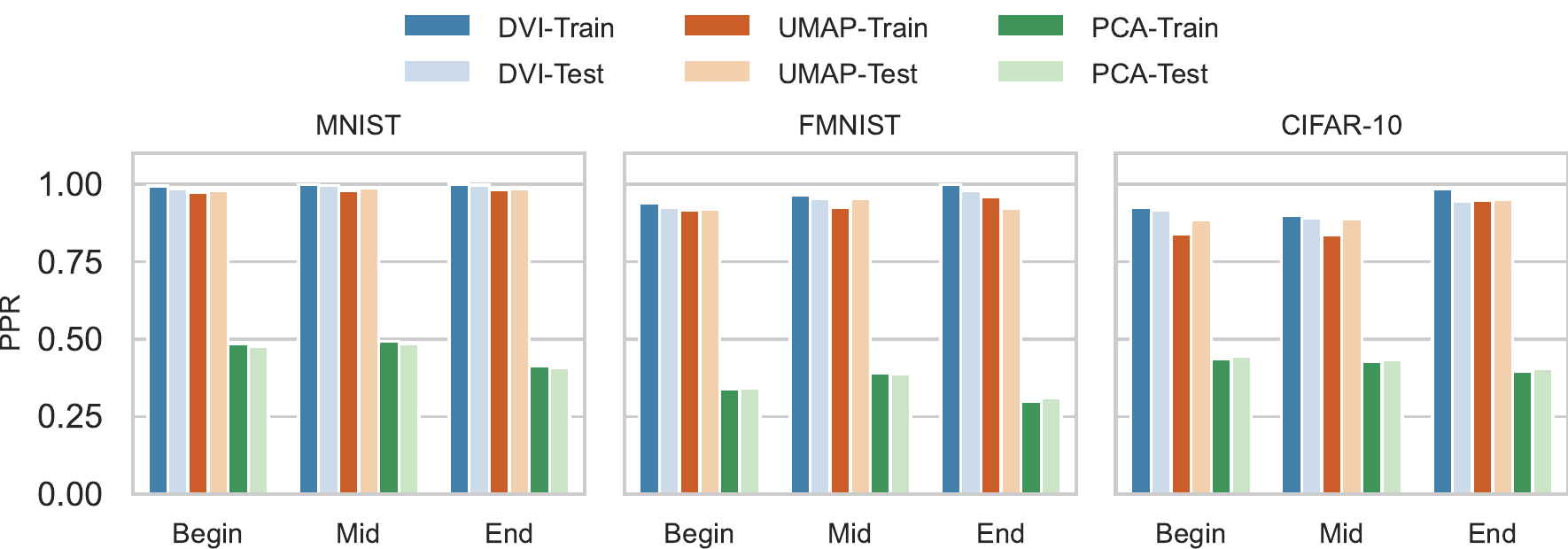}
    \caption{PPR between \tool, UMAP, and PCA}
    \label{fig:ppr}
    \vspace{-5pt}
\end{figure}

\begin{figure}[t]
    \centering
    \includegraphics[width=0.47\textwidth]{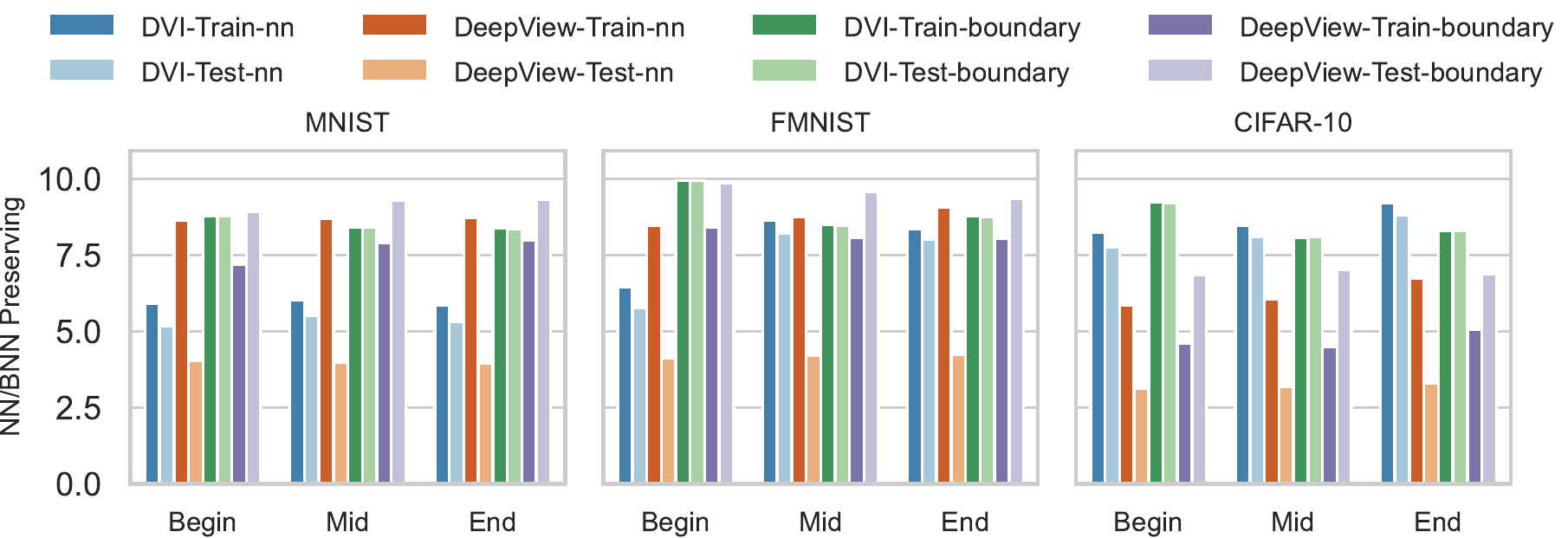}
    \caption{$k$-(Boundary/non-Boundary) Neighbour Preserving between \tool and DeepView ($k$=15)}
    \label{fig:dv_nnbnn}
    \vspace{-5pt}
\end{figure}

\begin{figure}[t]
    \centering
    \includegraphics[width=0.47\textwidth]{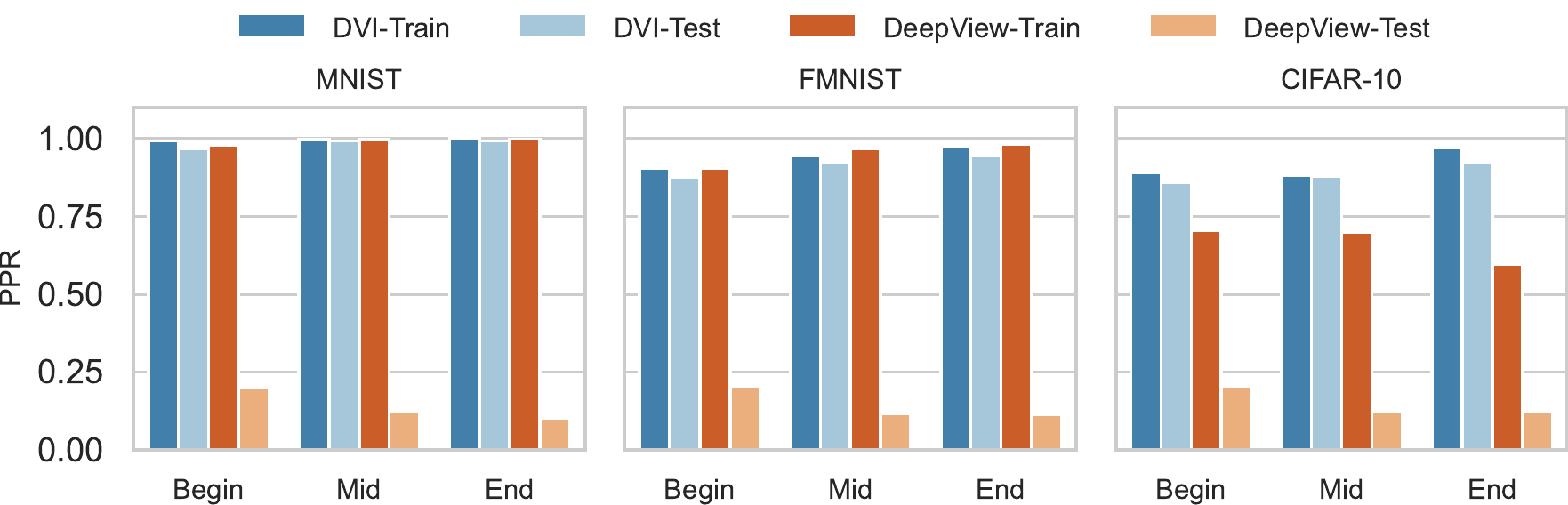}
    \caption{PPR between \tool and UMAP}
    \label{fig:dv_ppr}
    \vspace{-5pt}
\end{figure}

\begin{table}[t]
\scriptsize
\tabcolsep=0.18cm
\centering
\caption{Visualization Overhead (in seconds)}\label{tab:efficiency}
\vspace{-5pt}
\begin{tabular}{lcccc}
\toprule
\multicolumn{1}{l}{\textbf{Solution}}                                                                  & \multicolumn{1}{c}{\textbf{Overhead Type}} & \multicolumn{1}{c}{\textbf{CIFAR-10}} & \multicolumn{1}{c}{\textbf{MNIST}} & \multicolumn{1}{c}{\textbf{FMNIST}} \\ \hline

\multicolumn{1}{l}{\multirow{2}{*}{DVI}}                                                               & \multicolumn{1}{c}{Offline}                & \multicolumn{1}{c}{792.784}          & \multicolumn{1}{c}{914.921}        & \multicolumn{1}{c}{896.296}         \\ \cline{2-5}
\multicolumn{1}{l}{}                                                                                   & \multicolumn{1}{c}{Online}                 & \multicolumn{1}{c}{0.016}            & \multicolumn{1}{c}{0.010}          & \multicolumn{1}{c}{0.010}           \\ \hline

\multicolumn{1}{l}{\multirow{2}{*}{UMAP}}                                                              & \multicolumn{1}{c}{Offline}                & \multicolumn{1}{c}{50.170}           & \multicolumn{1}{c}{58.311}         & \multicolumn{1}{c}{58.748}          \\ \cline{2-5}
\multicolumn{1}{l}{}                                                                                   & \multicolumn{1}{c}{Online}                 & \multicolumn{1}{c}{1819.598}         & \multicolumn{1}{c}{2187.888}       & \multicolumn{1}{c}{2150.703}        \\ 
\hline
\multicolumn{1}{l}{\multirow{2}{*}{tSNE}}                                                              & \multicolumn{1}{c}{Offline}                & \multicolumn{1}{c}{207.757}          & \multicolumn{1}{c}{286.068}        & \multicolumn{1}{c}{282.725}         \\ \cline{2-5}
\multicolumn{1}{l}{}                                                                                   & \multicolumn{1}{c}{Online}                 & \multicolumn{1}{c}{/}                & \multicolumn{1}{c}{/}              & \multicolumn{1}{c}{/}               \\ \hline

\multicolumn{1}{l}{\multirow{2}{*}{PCA}}                                                               & \multicolumn{1}{c}{Offline}                & \multicolumn{1}{c}{0.803}            & \multicolumn{1}{c}{0.958}          & \multicolumn{1}{c}{0.951}           \\ \cline{2-5}
\multicolumn{1}{l}{}                                                                                   & \multicolumn{1}{c}{Online}                 & \multicolumn{1}{c}{0.035}            & \multicolumn{1}{c}{0.036}          & \multicolumn{1}{c}{0.035}       \\ \hline
\hline
\multicolumn{1}{l}{\multirow{2}{*}{\begin{tabular}[c]{@{}l@{}}DVI \\ (1000 samples)\end{tabular}}}     & \multicolumn{1}{c}{Offline}                & \multicolumn{1}{c}{19.801}           & \multicolumn{1}{c}{17.150}          & \multicolumn{1}{c}{18.896}          \\ \cline{2-5}
\multicolumn{1}{l}{}                                                                                   & \multicolumn{1}{c}{Online}                 & \multicolumn{1}{c}{0.004}            & \multicolumn{1}{c}{0.004}          & \multicolumn{1}{c}{0.004}           \\ \hline
\multicolumn{1}{l}{\multirow{2}{*}{\begin{tabular}[c]{@{}l@{}}DeepView\\ (1000 samples)\end{tabular}}} & \multicolumn{1}{c}{Offline}                & \multicolumn{1}{c}{1305.229}         & \multicolumn{1}{c}{506.839}        & \multicolumn{1}{c}{506.394}         \\ \cline{2-5}
\multicolumn{1}{l}{}                                                                                   & \multicolumn{1}{c}{Online}                 & \multicolumn{1}{c}{563.471}          & \multicolumn{1}{c}{204.473}        & \multicolumn{1}{c}{204.436}         \\
\bottomrule
\end{tabular}
\vspace{-5pt}
\end{table}

\noindent\textbf{Results (Spatial Property).}
Figure~\ref{fig:nn}, \ref{fig:boundary}, \ref{fig:ppr} and Table~\ref{tab:efficiency} show the performances of \tool and PCA, UMAP, and t-SNE on spatial properties on three datasets.
Given the space limit, we show the results with $k=15$, which shares similar performance when $k\in\{10, 20\}$ (see \cite{dvi}).
We report the results of three representative epochs, i.e.,
the 1st epoch (representing the beginning epoch),
the ($\frac{1+n}{2}$)th epoch (representing the middle epoch),
and the $n$th epoch (representing the final epoch).
We observe as follows:
\begin{itemize}[leftmargin=*]
  \item \textbf{PCA vs \tool}:
    PCA is a highly efficient solution (see Table~\ref{tab:efficiency}). 
    But its linear transformation has limitation, thus outperformed by \tool and UMAP (see Figure~\ref{fig:nn}, \ref{fig:boundary}, \ref{fig:ppr}).
  \item \textbf{t-SNE vs \tool}:
    On training dataset, t-SNE significantly outperforms all other approaches regarding the preserved neighbours after projection.
    However, it cannot
    (1) generalize the projection to any unseen samples and
    (2) inverse-project a 2-dimensional point back to feature vector space.
    Moreover, t-SNE fails to preserve boundary neighbours as \tool and UMAP (see Figure~\ref{fig:boundary}).
  \item \textbf{UMAP vs \tool}:
    UMAP has comparable performance with \tool on the neighbour-preserving projection and prediction-preserving inverse-projection (as showed in Figure~\ref{fig:nn} and Figure~\ref{fig:ppr}).
    However, even trained with boundary samples, UMAP is largely outperformed by \tool regarding the boundary-neighbour preserving projection.
    Noteworthy, UMAP takes a much larger runtime overhead than \tool when inverse-projecting the low-dimensional points to the feature space ($\sim$16.8s for UMAP vs $\sim$0.002s for \tool, see Table~\ref{tab:efficiency}).
  \item \textbf{DeepView vs \tool}:
    Regardless of the limited scalability of DeepView,
    DeepView is outperformed by \tool regarding:
    \begin{enumerate}
      \item DeepView is more likely to overfit the training dataset, thus its preserved neighbours on the test set is much less than that on the training set (see Figure~\ref{fig:dv_nnbnn}).
      \item DeepView can hardly preserve the prediction results after projection and inverse-projection (see Figure~\ref{fig:dv_ppr}).
    \end{enumerate}
\end{itemize}

\noindent\textbf{Results (Temporal).}
We compares the $temporal_{pv}$ value on
(1) UMAP trained with transfer learning (denoted as UMAP-T);
(2) DVI trained with transfer learning but without temporal loss (denoted as DVI-T); and
(3) DVI (denoted as DVI);
The results are shown in on Table~\ref{tab:temporal-neighbour-preserving}.
Overall, DVI surpasses UMAP-T and DVI-T regarding the temporal continuity.

\begin{table}[t]
\scriptsize
\tabcolsep=0.18cm
\centering
\caption{Temporal Results, i.e., $temporal_{pv}$ value ($k$=15)}\label{tab:temporal-neighbour-preserving}
\vspace{-5pt}
\begin{tabular}{ccccccc}
\toprule
\multicolumn{1}{l}{\multirow{2}{*}{\textbf{Solution}}} & \multicolumn{2}{c}{\textbf{CIFAR-10}} & \multicolumn{2}{c}{\textbf{MNIST}} & \multicolumn{2}{c}{\textbf{FMNIST}} \\ \cline{2-7}
\multicolumn{1}{l}{}                                   & \textbf{train}    & \textbf{test}     & \textbf{train}   & \textbf{test}    & \textbf{train}    & \textbf{test}    \\ \hline
UMAP-T                                                   & -0.453            & -0.448            & -0.581           & -0.578           & -0.622            & -0.613           \\ \hline
DVI-T                                                    & -0.442            & -0.460            & -0.463           & -0.466           & -0.291            & -0.286           \\ \hline
\textbf{DVI}                                             & \textbf{-0.463}   & \textbf{-0.498}   & \textbf{-0.609}  & \textbf{-0.611}  & \textbf{-0.626}   & \textbf{-0.632}  \\
\bottomrule
\end{tabular}
\end{table}

\noindent\textbf{Runtime Efficiency.}
Table~\ref{tab:efficiency} shows the runtime efficiency of all the solutions.
In Table~\ref{tab:efficiency}, the offline overhead is the time spent on training the visualization model;
the online overhead is the time spent on visualizing a new sample.
Overall, \tool takes more time to train the encoder and decoder, while it is very efficient to visualize the runtime new data.
In contrast, UMAP is efficient to train but takes considerable time to inverse-project the low-dimensional points back to representation vector space.
Moreover, \tool outperforms DeepView in both the offline and online efficiency.

\begin{figure}[h]
  \centering
  \begin{subfigure}[b]{0.13\textwidth}
    \includegraphics[width=\textwidth]{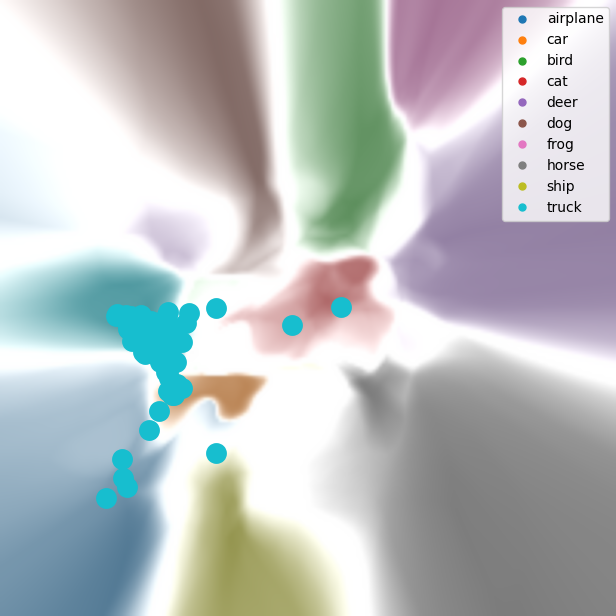}
    \caption{Random}\label{fig:al_random}
  \end{subfigure}
  ~
  \begin{subfigure}[b]{0.13\textwidth}
    \includegraphics[width=\textwidth]{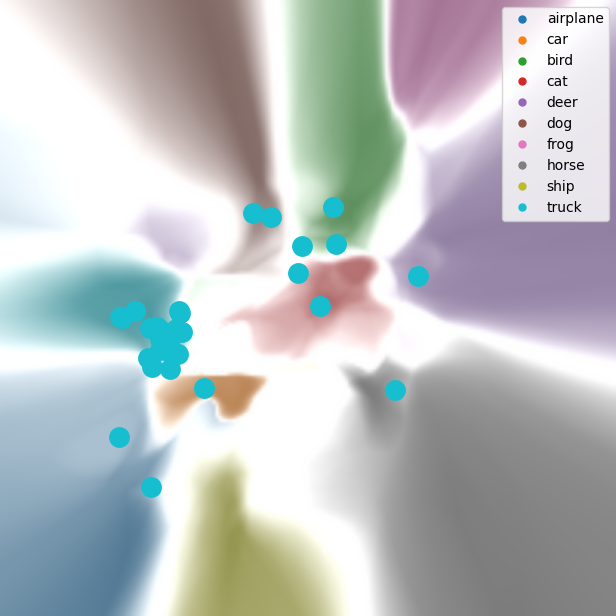}
    \caption{Coreset}\label{fig:al_coreset}
  \end{subfigure}
  ~
  \begin{subfigure}[b]{0.13\textwidth}
    \includegraphics[width=\textwidth]{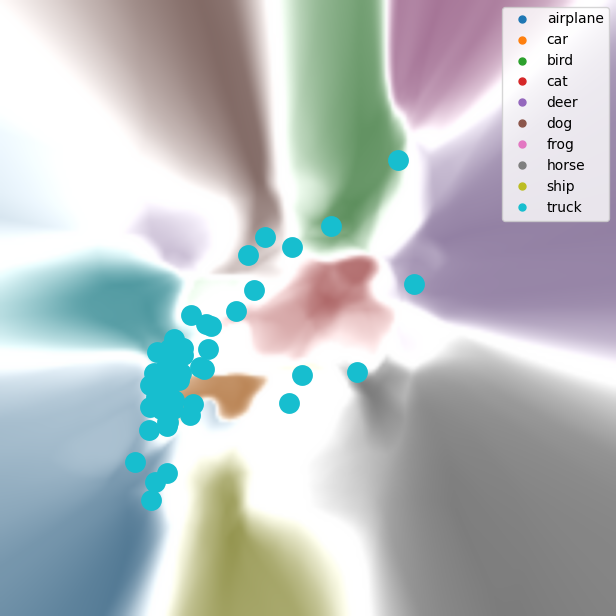}
    \caption{LL4AL}\label{fig:a_ll4al}
  \end{subfigure}

  \vspace{-5pt}
  \caption{Visualized active learning (sampling) strategies}\label{fig:al}
  \vspace{-15pt}
\end{figure}

\section{Case Study}\label{sec:case-study}

In this section, we introduce two case studies showing how \tool can support noise (hard) sample detection and active learning strategy comparison.
Readers can refer to more case studies in \cite{dvi}.

\subsubsection{Noise (Hard) Sample Detection}

\begin{figure}[h]
  \centering
  \begin{subfigure}[b]{0.08\textwidth}
    \includegraphics[width=\textwidth]{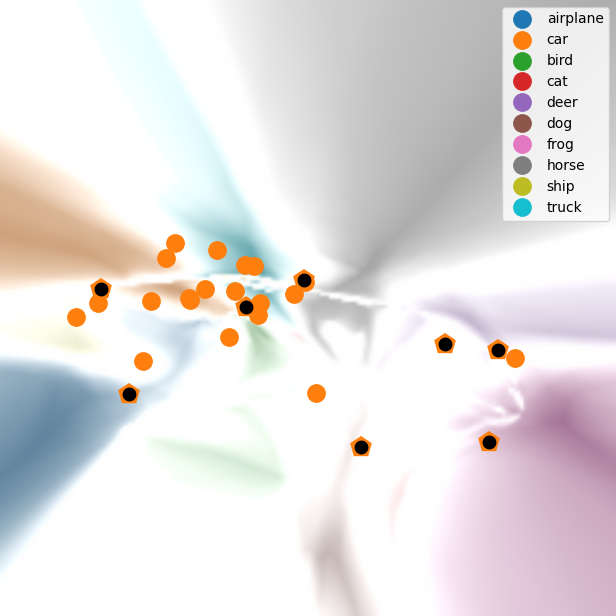}
    \caption{E10, 60.42\%}\label{fig:noise1}
  \end{subfigure}
  ~
  \begin{subfigure}[b]{0.08\textwidth}
    \includegraphics[width=\textwidth]{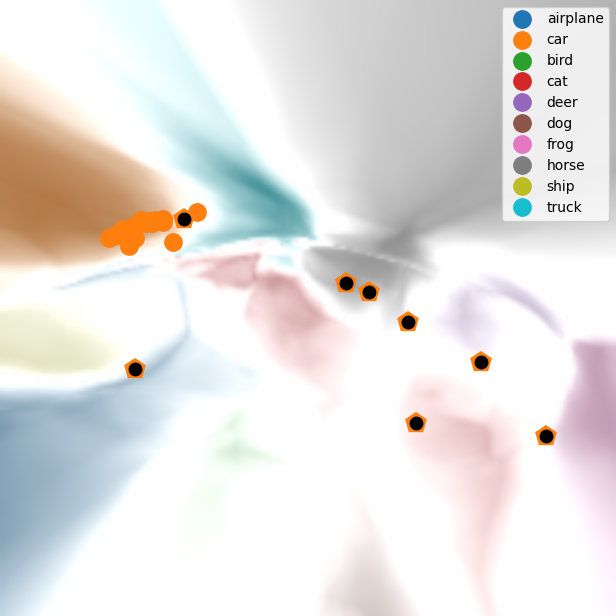}
    \caption{E50, 80.83\%}\label{fig:noise2}
  \end{subfigure}
  ~
  \begin{subfigure}[b]{0.08\textwidth}
    \includegraphics[width=\textwidth]{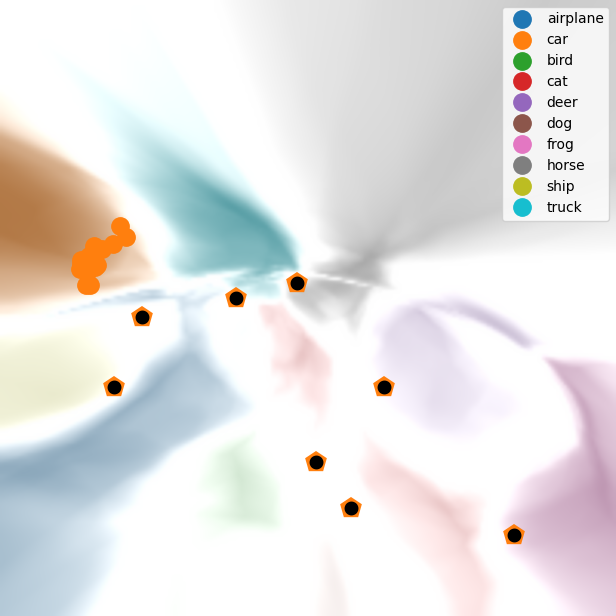}
    \caption{E100, 83.59\%}\label{fig:noise3}
  \end{subfigure}
  ~
  \begin{subfigure}[b]{0.08\textwidth}
    \includegraphics[width=\textwidth]{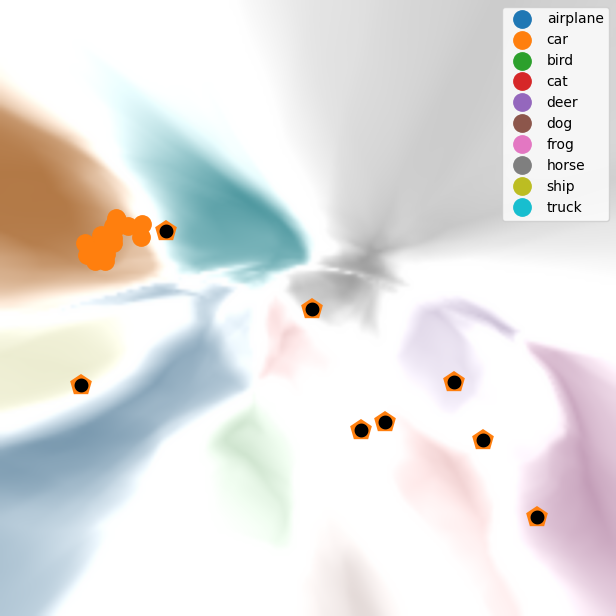}
    \caption{E150, 87.78\%}\label{fig:noise4}
  \end{subfigure}
  ~
  \begin{subfigure}[b]{0.08\textwidth}
    \includegraphics[width=\textwidth]{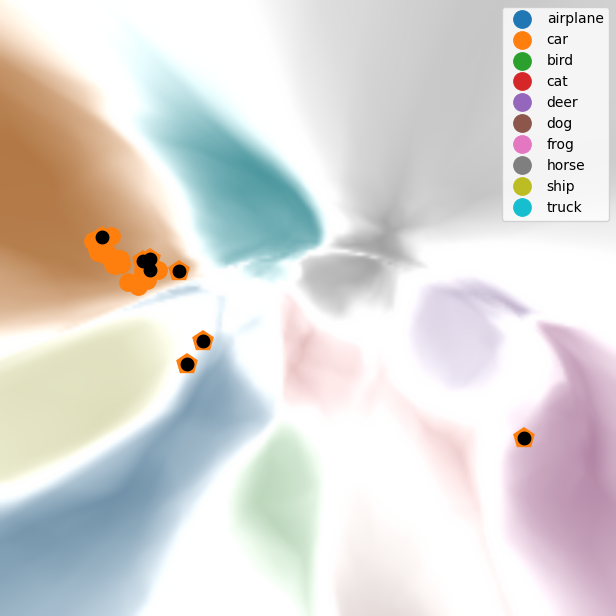}
    \caption{E200, 88.44\%}\label{fig:noise5}
  \end{subfigure}
  \vspace{-5pt}
  \caption{Visualized Training Process With Noise Data (epoch number and training accuracy)}\label{fig:noise}
\end{figure}

We generate symmetric noise by flipping the label of 10\% of the CIFAR-10 samples to train a classifier.
Figure~\ref{fig:noise} shows the process of how clean/noisy sample embeddings are learned during training.
For clarify, we show the dynamics of representative clean samples (orange dots) and noisy samples (orange dots tainted with a black core).
Comparing to the clean samples smoothly pulled into their color-aligned territory in the first few epochs,
noisy samples show ``reluctance'' to be pulled (i.e., learned).
Those ``hard'' samples continue to stay in their ``original'' territory in early-mid epochs, 
but some are forcefully pulled into their ``expected'' territory in late epoches.

By searching and pinpointing the interested samples and tracking their movements,
\tool can further allow users to zoom in to a local region and check the sample details including labels and appearances,
which serve as a potential model debugging facility.


\subsubsection{Active Learning Strategies Comparison}

Active learning algorithms sample the most informative unlabelled samples to retrain the classifier.
Various algorithms samples data regarding their diversity and uncertainty (i.e., how unconfident the classifier predict the samples).
Figure~\ref{fig:al} compares the new sampled data by different active learning algorithms on the same classification landscape.
We select Core-set \cite{sener2017active} and LL4AL \cite{yoo2019learning} as diversity and uncertainty based methods in this study.
Comparing to random (dots concentrated in color-aligned territory),
core-set selects samples that are more evenly distributed in whole landscape and LL4AL selects samples that lie closer to decision boundaries,
confirming the effectiveness of the two strategies.
Further investigation based on 
\tool allows users to inspect how those new selected samples are trained (i.e., pulled) and how they can influence the classification landscape in the subsequent epochs.



\break
\begin{small}
\bibliography{aaai22.bbl}
\end{small}

\section*{Acknowledgements}
We thank anonymous reviewers for their valuable input to improve our work. 
This work was supported in part by the Minister of Education, Singapore (No. MOET32020-0004, No. T2EP20120-0019 and No. T1-251RES1901), 
the National Research Foundation Singapore through its National Satellite of Excellence in Trustworthy Software Systems (NSOE-TSS)
office (Award Number: NSOE-TSS2019-05).

\end{document}